\pdfoutput=1
\documentclass[11pt]{article}

\usepackage[]{acl}
\usepackage{times}
\usepackage{latexsym}
\usepackage{amsfonts}
\usepackage{amsmath}
\usepackage{amssymb}
\usepackage{multicol}
\usepackage{multirow}
\usepackage{xspace}
\usepackage{booktabs}
\usepackage{bbding}
\usepackage{array}
\usepackage{threeparttable}
\usepackage{tcolorbox}
\usepackage{enumitem}
\usepackage{xcolor,colortbl}

\newcommand{\tabincell}[2]{
\begin{tabular}{@{}#1@{}}#2\end{tabular}
}

\usepackage{todonotes}

\usepackage{cleveref}
\crefformat{section}{\S#2#1#3}
\crefformat{subsection}{\S#2#1#3}
\crefformat{subsubsection}{\S#2#1#3}
\crefrangeformat{section}{\S\S#3#1#4 to~#5#2#6}
\crefmultiformat{section}{\S\S#2#1#3}{ and~#2#1#3}{, #2#1#3}{ and~#2#1#3}
\crefmultiformat{subsection}{\S\S#2#1#3}{ and~#2#1#3}{, #2#1#3}{ and~#2#1#3}
\usepackage{refstyle}
\Crefformat{figure}{#2Fig.~#1#3}
\Crefmultiformat{figure}{Figs.~#2#1#3}{ and~#2#1#3}{, #2#1#3}{ and~#2#1#3}
\Crefformat{table}{#2Tab.~#1#3}
\Crefmultiformat{table}{Tabs.~#2#1#3}{ and~#2#1#3}{, #2#1#3}{ and~#2#1#3}
\Crefformat{appendix}{Appx.~\S#2#1#3}
\crefmultiformat{appendix}{Appx.~\S#2#1#3}{ and~#2#1#3}{, #2#1#3}{ and~#2#1#3}
\crefformat{algorithm}{Alg.~#2#1#3}

\definecolor{ballblue}{rgb}{0.13, 0.67, 0.8}
\definecolor{azure1}{rgb}{0.0, 0.5, 1.0}
\definecolor{uclablue}{rgb}{0.33, 0.41, 0.58}
\definecolor{ultramarine}{rgb}{0.07, 0.04, 0.56}
\definecolor{yaleblue}{rgb}{0.06, 0.3, 0.57}

\newcommand{\xhdr}[1]{\vspace{0.3em}\noindent{{\bf #1.}}}
\newcommand{\modelname}{\textsc{SuRE}\xspace}
\newcommand{\modelnamens}{\textsc{SuRE}}
\newcommand{\modelnameexplain}{(\underline{Su}mmarization as \underline{R}elation \underline{E}xtraction)\xspace}

\usepackage[T1]{fontenc}
\usepackage[utf8]{inputenc}

\usepackage{microtype}

\title{Summarization as Indirect Supervision for Relation Extraction}

\author{
Keming Lu$^{\dagger}$, I-Hung Hsu$^{\dagger}$, Wenxuan Zhou$^{\dagger}$, Mingyu Derek Ma$^{\ddagger}$ \and Muhao Chen$^{\dagger}$\\
$^{\dagger}$University of Southern California\\
$^{\ddagger}$University of California, Los Angeles \\ 
\texttt{\{keminglu,ihunghsu,zhouwenx,muhaoche\}@usc.edu};\; \texttt{{ma@cs.ucla.edu}}\\
}

\begin{document}
\maketitle
\begin{abstract}
Relation extraction (RE) models have been challenged by their reliance on training data with expensive annotations.
Considering that summarization tasks aim at acquiring concise expressions of synoptical information from the longer context, these tasks naturally align with the objective of RE, i.e., extracting a kind of synoptical information that describes the relation of entity mentions.
We present \modelname, which converts RE into a summarization formulation. 
\modelname leads to more precise and resource-efficient RE based on indirect supervision from summarization tasks.
To achieve this goal,
we develop sentence and relation conversion techniques that essentially bridge the formulation of summarization and RE tasks.
We also incorporate constraint decoding techniques with Trie scoring to further enhance summarization-based RE with robust inference.
Experiments on three RE datasets demonstrate the effectiveness of \modelname in both full-dataset and low-resource settings, showing that summarization is a promising source of indirect supervision signals to improve RE models.\footnote{Our code is public available at \url{https://github.com/luka-group/SuRE}}
\end{abstract}

\section{Introduction}
% Para1: Background. Importance of relation extraction

Relation extraction (RE) aims at extracting relations between entity mentions from their textual context. 
For example, given a sentence \emph{``Steve Jobs is the founder of Apple''}, an RE model would identify the relation \emph{``founded''} between mentioned entities \emph{``Steve Jobs''} and \emph{``Apple''}. 
RE is a fundamental natural language understanding task and is also the essential step of structural knowledge acquisition for constructing knowledge bases. Hence, advanced RE models is crucial for various knowledge-driven downstream tasks, such as dialogue system~\cite{liu-etal-2018-knowledge,zhao-etal-2020-knowledge-grounded}, narrative prediction~\cite{chen2019incorporating}, and question answering ~\cite{yasunaga-etal-2021-qa,hao-etal-2017-end}. 

% Para2: Existing works in relation extraction
Given sentences with detected pairs of entity mentions, most recent studies formulate RE as multi-class classification~\cite{zhou2021improved,yamada-etal-2020-luke,baldini-soares-etal-2019-matching}. 
Models presented in these studies employ pretrained language models (PLM) equipped with classification heads and are finetuned with a cross-entropy loss.
Although such methods have achieved enhanced performances on several benchmarks~\cite{zhang-etal-2017-position,stoica2021re,alt2020tacred}, they fall short of capturing the semantic meanings of the relations.
This shortage hinders PLMs from effectively matching the sentential context with the relations that are merely converted as logits. %for classification.
On the other hand, obtaining high-quality annotations for RE is often costly due to the difficulty for annotators to recognize and mutually agree on such structural information.
This represents another challenge for RE models that %rely on large-scale relation annotations to be applied in real-word applications.
have relied on direct supervision from sufficient end-task training data.
Existing literature finds that classification models have drastically degraded performance under low-resource scenarios~\cite{sainz-etal-2021-label},
%Therefore,
showing that label efficiency is a vital issue when adopting prior methods in real application scenarios.
%Indirect supervision allows the use of additional supervision signals that is not specific to RE, which can potentially combat this issue.
%Thus, in this work, we aim at discovering effective indirect supervision signals that can help extracting relations, especially when direct RE annotations are not adequately available.
To combat this issue, we aim at investigating an indirectly supervised method for RE, which allows the use additional supervision signals that are not specific to RE 
%we aims at discovering a new form of indirect supervision signals that can effectively enhance RE, especially when direct RE annotations are not adequately available.
without solely relying on direct RE annotations.

\begin{figure*}[t]
    \centering
    \includegraphics[width=0.98\linewidth]{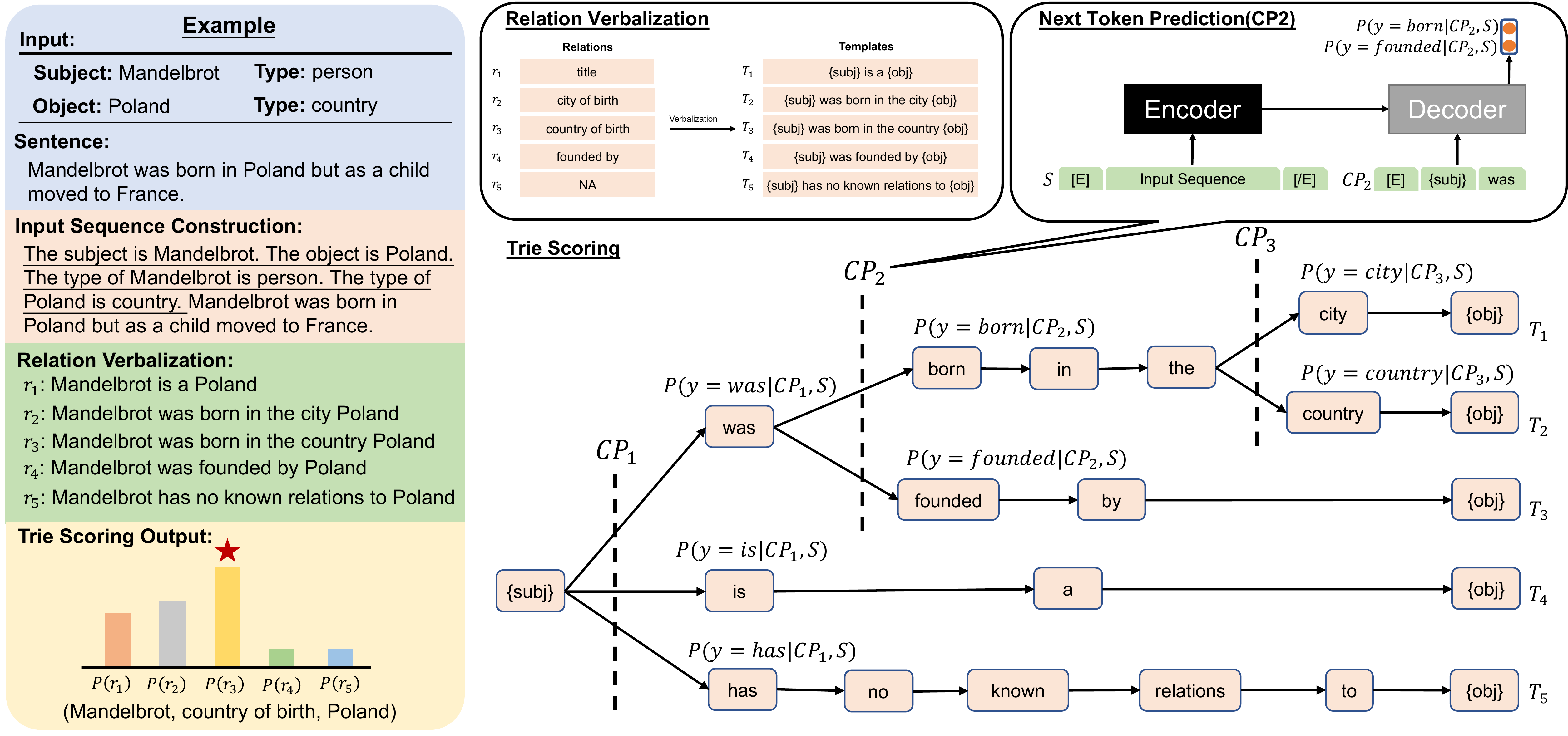}
    % \vspace{-0.4em}
    \caption{Overview of \modelname inference with an example. \modelname constructs input sequences and verbalizes candidate relations with semantic templates (\underline{Relation Verbalization} subfigure). Then we use a summarization model to find the best prediction by Trie scoring technique (\underline{Trie scoring} subfigure). This model is pretrained on summarization tasks and then simply finetuned with input sequence target verbalized ground-truth relation.
    For each common prefix (CP), we will calculate the probability of each relation candidate, which is obtained using a trained summarization model, as shown in the \underline{Next Token Prediction} subfigure. The ``\{subj\}'' and ``\{obj\}'' are two placeholders representing subject and object entity in each sample. 
    }
    \label{fig:model}
    % \vspace{-0.9em}
\end{figure*}

% Para3: Relation extraction as summarization
This study proposes \modelname~\modelnameexplain, which reformulates and addresses RE as a summarization task.\footnote{In this paper, we specifically consider abstractive summarization instead of extractive summarization. Since relation labels are often not directly expressed in sentences, extractive summarization does not always support the inference of RE.} Summarization seeks to acquire concise expressions of synoptical information from longer context~\cite{el2021automatic}, which aligns well with the objective of RE if we consider the relation between entities as one aspect of synoptical information in the sentential context. Such an affinity of task objectives naturally motivates us to leverage indirect supervision from summarization tasks to improve RE models. %Summarization models can also effectively capture semantics of relations.
In comparison to a multi-class classifier, summarizing the relation information also allows generating a semantically rich representation of the relation.
Furthermore, unlike existing RE models that rely on costly manual annotations of structural information on training sentences, summarization tasks allow training with considerably richer and unannotated parallel text corpora.\footnote{Particularly, summarization corpora are constructed in the scales from hundred thousands \cite{nallapati-etal-2016-abstractive,narayan-etal-2018-dont} to million scales \cite{yin-etal-2021-docnli}, and may be rapidly augmented in large scales from easy-to-consume data sources (e.g., community Q\&A platforms \cite{mishra-etal-2021-looking} and scientific paper abstracts \cite{cohan-etal-2018-discourse}).}
Hence, summarization tasks can bring in abundant indirect supervision signals, and can potentially lead to label-efficient models under scenarios without much task-specific annotation.

% Para4: Our works

% summarize our work
\Cref{fig:model} illustrates the structure of \modelname. Specifically, \modelname transforms RE to a summarization task with relation and sentence conversion techniques (\Cref{ssec:conversion}), and applies constrained inference for relation prediction (\Cref{ssec:inference}).
We deploy an entity information verbalization technique to highlight the sentential contexts with entity information,
and verbalize relations into template-style short summaries. In that way, the converted inputs and outputs of RE naturally suit a summarization model.
Then, we adapt a summarization model to the RE task by finetuning it on the converted RE data (\Cref{ssec:training}). 
During inference, a Trie scoring technique is designed to infer the relations (\Cref{ssec:inference}).
In this way, 
\modelname fully utilizes indirect supervision from summarization, %which benefits RE in low-resource setting and provide more precise RE with adequate data.
%leading to more precise and label-efficient RE even in low-resource scenarios.
allowing a precise RE model to be obtained even in low-resource scenarios.
%Besides, it can exploit relation semantics via our verbalization technique.

% Contributions
The contributions of this work are two-folds.
First, to the best of our knowledge, this is the first study on using indirect supervision from summarization for RE. Since the objective of summarization naturally aligns with RE, it allows precise RE models to be trained without solely relying on direct task annotations, 
and benefits with robust RE under low-resource scenarios.
Second, we investigate input conversion techniques that %essentially
effectively bridge the formulation of summarization and RE tasks,
as well as constraint techniques that further enhance the inference of summarization-based RE.
Our contributions are verified with experiments %conducted 
on three widely used sentence-level RE datasets, TACRED, TACREV, and SemEval, as well as three low-resource settings of TACRED. We observe that \modelname outperforms various baselines, especially in the low-resource setting with 10\% TACRED training data. \modelname also achieves SOTA performance with 75.1\% and 83.5\% in micro F1 on TACRED and TACREV, respectively. We also perform comprehensive ablation studies to show the effectiveness of indirect supervision from summarization and the best options of input conversion for \modelname.

\section{Related Work}\label{sec:related}
% \vspace{-0.3em}
%We introduce three lines of related works.

% \vspace{-0.3em}
\xhdr{Relation extraction}
Recent studies on (sentence-level) RE typically formulate the task as multi-class classification tasks by finetuning pretrained language models~\cite{wu2019enriching,hsu2021discourse,lyu-chen-2021-relation} or developing pretraining objectives for RE~\cite{baldini-soares-etal-2019-matching,peng-etal-2020-learning}.
For example, 
\citet{wu2019enriching} 
enrich the contextual representation of the PLM with marked-out subject and object entity mentions.
% \citet{baldini-soares-etal-2019-matching}
% propose to match similar sentential contexts to provide distant supervision for extracting each type of relation.
\citet{lyu-chen-2021-relation} propose a model-agnostic paradigm that introduces mutual restrictions of relations and entity types into relation classifiers.
On the basis of PLMs, some studies further improve RE with external knowledge from knowledge bases (KBs)~\cite{yamada-etal-2020-luke,peters-etal-2019-knowledge,zhang-etal-2019-ernie}.
% For example, \citet{yamada-etal-2020-luke} develop an improved contextualized entity representation by modifying the Transformer PLM with an entity-aware self-attention mechanism. 
More studies have been introduced to improve entity pair representations and classifiers for RE such that we cannot exhaust them in this short summary. We refer readers to the recent benchmarking study \cite{zhou2021improved}.

Another threads of recent effort in RE introduce several reformulations of RE with prompt learning~\cite{han2021ptr,chen2021knowprompt}.
% and indirect supervision~\cite{joshi-etal-2020-spanbert,sainz-etal-2021-label,levy-etal-2017-zero,cohen2020relation}. 
Specifically,
% \citet{joshi-etal-2020-spanbert} present a span-prediction based model for RE and show that it outperforms former multi-class classification models.
\citet{han2021ptr} propose prompt tuning methods for RE by applying logic rules to construct hierarchical prompts.
\citet{chen2021knowprompt} leverage prompt tuning for RE by injecting semantics of relations and entity types. 
%
% Instead of taking indirect supervision from masked language modeling or NLI, our work is the first to leverage indirect supervision from summarization, which has a task objective that naturally aligns with RE.
Instead of leveraging pretrained masked language models, 
we use generative approaches to solve RE.

\xhdr{Indirect supervision} Indirect supervision \cite{he-etal-2021-foreseeing} methods often modify the training and inference processes on a task into a different formation, hence allowing the use of additional supervision signals that is not specific to this task.
% In addition to the aforementioned works,
\citet{levy-etal-2017-zero} show that RE can be addressed as answering reading comprehension questions and improved by the training process of a machine reading comprehensive task. 
Similarly,
% \citet{li-etal-2019-entity} formulate the joint entity and relation extraction tasks as a multi-turn question answering problem.
\citet{wu-etal-2020-corefqa} also employ QA data to improve model generalization abilities in coreference resolution.
\citet{yin2020universal} propose a few-shot NLI-based framework to address different tasks, such as question answering and coreference resolution. 
\citet{li2022ultra} further improve this strategy by incorporating NLI with learning-to-rank, leading to a robust system for ultra-fine entity typing~\cite{choi2018ultra}.
Similar idea of leveraging NLI as indirect supervision signal is applied by \citet{sainz-etal-2021-label}, which focuses on low-resource RE task.
As discussed, the objective of RE aligns well with that of a summarization task. While there is no prior study that investigates indirect supervision from summarization, this is exactly the focus of our study.

\xhdr{Generative approaches for discriminative tasks}
Formulating discriminative tasks as
generation tasks can be an efficient way to guide PLMs to leverage semantics of decision labels~\cite{huang-etal-2021-document,hsu2021degree,acl2022xgear,yuan-etal-2022-generative}. Instead of predicting classification logits, a common paradigm for these models is to represent the class as a concise structure
and employ controlled decoding for generation. 
Several studies \cite{zeng-etal-2018-extracting, zeng2020copymtl, ye2020contrastive, cao-ananiadou-2021-generativere-incorporating} use sequence-to-sequence-based models to generate relations written in a triplet format.
\citet{paolini2020structured} incorporate many structured prediction tasks, including RE, into machine translation. \citet{huguet-cabot-navigli-2021-rebel-relation} simplify RE as expressing relations as a sequence of text to perform end-to-end generation of relations. 
These works mostly formulate RE as text-to-structure learning instead of generating natural language sentences that is a more natural target to exploit the power of pretrained generative models~\cite{hsu2021degree}.
Additionally, they do not include indirect supervision of summarization, which is naturally close to the objective of RE and has the potential to benefit RE performance. 

\section{Method}
In this section, we describe \modelname, a model for addressing RE with summarization.
We introduce preliminaries (\Cref{ssec:preliminaries}), how RE data are converted to suit summarization tasks (\Cref{ssec:conversion}), training (\Cref{ssec:training}), and constrained inference of \modelname (\Cref{ssec:inference}).

\subsection{Preliminaries}\label{ssec:preliminaries}

% \vspace{-0.3em}
\xhdr{Problem Definition}
The input to the sentence-level RE is a sentence $s$ with entity mentions $e_1$ and $e_2$~\footnote{An entity mention is presented as an \emph{entity name} in text, and is structured as a \emph{mention span} with position information.}
, where their auxiliary entity type information $t_1, t_2$ is given.
An RE model aims at inferring the relation $r$ between the subject and object entities $e_1$ and $e_2$ from a set of candidate relations $\mathcal{R}= \mathcal{R}_P\cup \{\emptyset\}$, which include positive relations $\mathcal{R}_P$ and a Not-Available (NA) relation $\emptyset$. We also involve type-related candidate relations $\mathcal{R}(t_1, t_2)$, which is a subset of $\mathcal{R}$ which has specific types of head and tail entities.

\xhdr{Overview} \Cref{fig:model} demonstrates the overview of \modelname. 
The summarization task takes a context as the input sequence and a summary target is expected to be generated.
To formulate RE as summarization, we first need to hint the summarization model which entity pair is targeted for summarization. To do so, we process the input sentence such that entity mentions and their type information will be highlighted (\Cref{ssec:conversion}).
We explore existing entity marking tricks \cite{zhou2021improved} and also develop entity information verbalization technique that directly augments entity information as part of the context.
The processed sentence will then be fed into \modelname.
The summary targets for \modelname is created via verbalizing existing RE labels to templates, such as the \underline{Relation Verbalization} subfigure in \Cref{fig:model}. In the training process (\Cref{ssec:training}), 
\modelname uses pretrained summarization models as a start point, and finetunes them with processed sentences as the input and verbalized relation descriptions as the targets.
During inference, %to decide the predicted relation, we develop a Trie scoring technique on \modelname to rank the probabilities for each relation candidate
we incorporate several constrained inference techniques to help \modelname decide the inferred relation
(\Cref{ssec:inference}).

\subsection{
Relation and Sentence Conversion
}\label{ssec:conversion}

Summarization takes text sequences as inputs and outputs. We hereby describe the input sequence construction and relation verbalization, representing two essential techniques for converting RE data to suit the summarization task.

\xhdr{Input sequence construction} 
Relation extraction focuses on analyzing the interaction between two specific entities, so we need to further process source sentences so that additional information can be involved and captured by summarization models.
\modelname explores a series of sentence processing techniques that highlight and incorporate entity information, aiming for identifying a technique that suits the summarization task well. Entity information includes entity names, types, and spans, which is useful for inferring the relation. 
We explore with two strategies for processing the source sentence.
\begin{itemize}[leftmargin=1em]
    \setlength\itemsep{0em}
    \item \textbf{Entity typed marker.} Various entity marking techniques are widely adopted in previous multi-class classification RE systems~\cite{zhang-etal-2017-position,zhang-etal-2019-ernie,wang2021k,zhou2021document,zhong-chen-2021-frustratingly,zhou2021improved}. We list all the techniques in Appx.~\Cref{tab:entity-information-injection-techniques}. Our preliminary experiments (Appx.~\Cref{tab:result-input-formulation-bart-large-cnn}) find that the following typed entity marker technique with punctuation works the best for \modelname among these marking methods (inserted typed markers are in {\color{yaleblue}blue}, while the original text is in black):
    \begin{tcolorbox}
    %%%\vspace{-0.5em}
    \small
         \textit{\color{yaleblue}@ * person *} Mandelbrot \textit{\color{yaleblue}@} was born in Poland but as a child moved to \textit{\color{yaleblue}\# $\wedge$ country $\wedge$} France \textit{\color{yaleblue}\#}.
    %%%\vspace{-0.5em}
    \end{tcolorbox}
    \item \textbf{
    Entity information verbalization.
    %Sentence rewriting
    } We develop a simple sentence rewriting technique that directly describes entity information as an augmented part of the linguistic context (in {\color{yaleblue}blue}):
    \begin{tcolorbox}
    %%%\vspace{-0.5em}
    \small
        \textit{\color{yaleblue}The subject entity is Mandelbrot. The object entity is France. The type of Mandelbrot is person. The type of France is country.} Mandelbrot was born in Poland but as a child moved to France.
    %%%\vspace{-0.5em}
    \end{tcolorbox}
    Although this technique cannot encode entity span information, it keeps the input data close to natural language instead of adding special tokens. This aligns well with the indirect supervision from summarization. Thus, it shows better performance to the entity typed marker technique, as shown in the ablation study (\Cref{ssec:input_formulation}).
\end{itemize}
%These two techniques can be easily combined by adding markers to the sentence first and then applying entity information verbalization (\Cref{ssec:additional_input_conversion}).
We hereby list all input conversion techniques we experiment in this work in \Cref{tab:entity-information-injection-techniques}. 
\Cref{tab:result-input-formulation-bart-large-cnn} shows additional results on the  \emph{bart-large-cnn} model, which provides the same conclusion as results on \emph{pegasus-large}.
We also compare this mixing technique in the ablation study (\Cref{ssec:input_formulation}) and find it achieves the best performance in the full training setting.

\xhdr{Relation verbalization} The target of summarization is verbalized by a set of simple \emph{semantic templates}, as shown in the \underline{Relation Verbalization} subfigure of \Cref{fig:model}.
Each template contains \emph{\{subj\}} and \emph{\{obj\}} placeholders to be filled with subject and object entity mentions in the sentence.
The templates seek to form short summaries that describe the relations between two entities, and will be used for models' training and inference.
Semantic templates are also leveraged in \citet{sainz-etal-2021-label}, where the templates are used as hypotheses for NLI-based RE. 
However, their templates are specifically designed for NLI. We adapt minimal additional updates to their templates so the templates can better fit summarization and less human effort is involved.
We let the subject entities always appear in the head of sentences while the objects are in the tail. For example, ``\emph{org:parents}'' relation is verbalized with templates ``\emph{\{subj\} has the parent company \{obj\}}''. Detailed semantic templates are demonstrates in Appx.~\Cref{tab:semantic1-templates} and ~\Cref{tab:semantic1-templates-semeval}. 
Notice that semantic meaning of a relation can be verbalized in various ways, so we also construct alternative semantic templates and discuss how different templates influence model performance in \Cref{ssec:template_design}.

For comparison, we also experiment with \emph{structural templates} that are widely used in existing sequence-to-structure methods~\cite{zeng-etal-2018-extracting,zeng2020copymtl}.
As listed in Appx.~\Cref{tab:structural-templates}, these templates directly concatenate entity names and relations, which are shown by our ablation study (\Cref{ssec:template_design}) to be less effective than the semantic templates.

\xhdr{Discussion}
\modelname requires manually designed templates of relations for both training and inference. To minimize manual effort and give a fair comparison to prior work, we adopt the same relation verbalization templates from \citet{sainz-etal-2021-label}. They restrict the influence of human effort by limiting
the time for creating the templates and build 2 templates in average for each relation. For simplicity, we adopt one template for each relation from their templates, which suggests \modelname will need less manual effort for template design.

\subsection{Training Process}\label{ssec:training}

The aforementioned rewriting and verbalization techniques (\Cref{ssec:conversion}) highlight the sentential contexts with entity information, and convert the extraction as summary.
Hence, the converted inputs and outputs of RE naturally suit the summarization task. 
This allows us to train a summarization model first using large parallel training corpora for abstractive summarization such as XSum~\cite{narayan-etal-2018-dont} or CNN/Dailymail~\cite{hermann2015teaching}, and further adapt it to learn to ``summarize'' relation. 
In our experiments, \modelname adopts checkpoints of pretrained generative models~\cite{lewis-etal-2020-bart, zhang2020pegasus} that are pretrained on summarization tasks as starting points. Then, we follow the same finetuning process of seq-to-seq training with the cross entropy loss to finetune the model on converted RE data.
In this way, \modelname can leverage indirect supervision obtained from the summarization task to enhance RE.

\subsection{Inference}\label{ssec:inference}
The inference process of \modelname involves first applying Trie scoring to rank the possibility of each relations, and setting entity type constraints. %Based on the score, we will select the best suited relations for the given entity pair.
The score is further calibrated to make selective predictions between known and NA relations.

\xhdr{Trie scoring} 
Summarization models employ beam search techniques to generate sequential outputs~\cite{zhang2020pegasus}, while RE seeks to find out the relation described the input. To support relation prediction using a summarization model, we develop an inference method that will rank each relation candidate by using the summarization model as proxies for scoring. Inspiring by the Trie constrained decoding \cite{decao2020autoregressive}, we develop a Trie scoring technique, allowing \emph{efficient ranking} for candidate relation verbalizations.
Instead of calculating the probability of whole %template sequences
relation templates for ranking, our method conducts a traverse on the Trie and estimates the probability of each relation candidate as a path probability on the Trie.

% Describe Trie
Given the set of tokenized templates of all candidate relations as $\mathcal{T}=\{T_i\}_{i=1}^{n_r}$, we build a Trie \cite{aoe1992efficient} by combining the prefixes of all templates, as an example in the \underline{Trie Scoring} subfigure in \Cref{fig:model}.
A path of a relation template can be described as a sequence of decision processes, which goes from the root to a leaf node. If we denote $\mathcal{N}^f$ as the set of forky nodes (the nodes with more than one child), then the probability of a path can be estimated by continued producting the probability of choosing a specific child in a forky node $n^f_i \in \mathcal{N}^f$.
Specifically, we denote the path from root to $n^f_i$ as $\textrm{\textrm{CP}}_i$, which is the common prefix for all templates in the sub-tree with $n^f_i$ as the root. For example, $\textrm{CP}_2=\textrm{``\{subj\} was''}$ in \Cref{fig:model} is the common prefix of templates $T_1$ and $T_2$.
If we denote the children of $n^f_i$ as a set $\mathbf{C}(n^f_i)$, the prediction probability of relation $r_i$ can be calculated by
\begin{equation*}
    p(r_i) = \Pi_{\forall n^f_j, c\in \mathbf{C}(n^f_j)} p(c\in T_i|\textrm{CP}_j, S).
\end{equation*}
$p(c\in T_i|\textrm{CP}_j, S)$ thereof is the probability for the model to generate the next token $c$ given the previous common prefix $\textrm{CP}_j$, and $c$ is %exactly the next node of $n_i^f$ on $T_i$. 
%selected from the next nodes of $n_i^f$.
selected from $C(n_i^f)$ on $T_i$.
This probability is calculated using a seq-to-seq summarization model with input sentence $S$ as encoder input and $\textrm{CP}_i$ as decoder input prefix, such as the illustration in the \underline{Next Token Prediction(CP2)} subfigure in \Cref{fig:model}.

\xhdr{Type constrained inference}
Type constrained inference emerges in many recent works~\cite{lyu-chen-2021-relation,sainz-etal-2021-label,cohen2020relation}. By constructing a type-relation mapping from the training set, models can only predict valid relations given entity types, which significantly shrinkages the size of the candidate relation set.
Type constrained inference can be easily incorporated into \modelname by pruning the templates of invalid entity types from Trie scoring.
The merit of type constrained inference will be discussed in \Cref{ssec:type_constrained_decoding}.

\xhdr{Calibration for NA relation}
Considering that it is not possible for a relation ontology to exhaust all possible relations between any entities, the inference of a trained RE model can naturally be exposed to many instances where not a positive relation from $\mathcal{R}_P$ is expressed.
Hence, it is particularly important to enforce the model to selectively make a decision between positive relations or predicting abstention.
This is realized by a calibration technique in \modelname, where a score threshold $s$ is set for NA and is calibrated as below:
\begin{equation*}
    \hat{r} = \left\{
    \begin{array}{cc}
    \arg\max_{r_i\in \mathcal{R}_P} p(r_i)\;,\;& p(\textrm{NA}) \leqslant s \\
    \textrm{NA}\;,\; & p(\textrm{NA}) > s
    \end{array}
    \right.
\end{equation*}
The best threshold $s$ is found on the development set and is used as a fixed threshold in inference.

\section{Experiments}

In this section, we present the experimental evaluation of \modelname for RE under both high- and low-resource setups (\Cref{ssec:experimental_setup}-\Cref{ssec:results}).
In addition, we also conduct comprehensive ablation studies to investigate the effectiveness of the incorporated techniques in \modelname (\Cref{ssec:ablation}).

\subsection{Experimental Setup}\label{ssec:experimental_setup}
\xhdr{Datasets} We conduct experiments on three widely used sentence-level RE benchmarks: SemEval 2010 Task 8 (SemEval; \citealt{hendrickx-etal-2010-semeval}), TACRED~\cite{zhang-etal-2017-position}, and TACRED-Revisited (TACREV; \citealt{alt2020tacred}). 

SemEval is an RE dataset which does not provide entity types, so we simply remove the processing of entity types in sentence conversion (\Cref{ssec:conversion}) to adapt \modelname on this dataset.
TACRED contains entity pairs drawn from the yearly TAC-KBP challenge. 
We list our templates for SemEval and TACRED in Appx. \Cref{tab:semantic1-templates-semeval} and  \Cref{tab:semantic1-templates}, respectively. 
TACREV relabeled develop and test sets of TACRED to correct mislabeled entity types and relations. 
TACREV shares the same templates with TACRED since they have exactly the same relations. 
Statistics of these datasets are showed in Appx. \Cref{tab:dataset-statistics}.
We report macro F1 on SemEval with the official grading script for this benchmark\footnote{The metric calculated by the script is the macro F1 on (9+1)-way classification taking directionality into account.}, and micro F1 on TACRED and TACREV to keep consistency with previous works~\cite{yamada-etal-2020-luke,zhou2021improved}.

\begin{table*}[t]
    \centering
    \begin{threeparttable}
    {
    \small
    \begin{tabular}{lccccccc}
    \toprule
    &\multirow{2}{*}{Dataset} & \multicolumn{4}{c}{TACRED} & \multirow{2}{*}{TACREV} & \multirow{2}{*}{SemEval} \\
    & & 1\% & 5\% & 10\% & 100\% &  &  \\
    \midrule
    \cellcolor{blue!10} & SpanBERT~\cite{joshi-etal-2020-spanbert} & 0.0\tnote{\ddag} & 28.8\tnote{\ddag} & 1.6\tnote{\ddag} & 70.8 & 78.0 & $--$ \\
    \cellcolor{blue!10}&KnowBERT~\cite{peters-etal-2019-knowledge} & $--$ & $--$ & $--$ & 71.5 & 79.3 & 89.1 \\
    \cellcolor{blue!10}&RoBERTa~\cite{wang2021k} & 7.7\tnote{\ddag} & 41.8\tnote{\ddag} & 55.1\tnote{\ddag} & 71.3 & $--$ & $--$ \\
    \cellcolor{blue!10}&R-BERT~\cite{wu2019enriching} & $--$ & $--$ & $--$ & 69.4 & $--$ & 89.3 \\
    \cellcolor{blue!10}&MTB~\cite{baldini-soares-etal-2019-matching} & $--$ & $--$ & $--$ & 71.5 &  $--$ & 89.5 \\
    \cellcolor{blue!10}&K-Adapter~\cite{wang2021k} & 13.8\tnote{\ddag} & 51.6\tnote{\ddag} & 56.0\tnote{\ddag} & 72.0 & $--$ & $--$ \\
    \cellcolor{blue!10}&LUKE~\cite{yamada-etal-2020-luke} & 17.0\tnote{\ddag} & 51.6\tnote{\ddag} & 60.6\tnote{\ddag} & 72.7 & 80.6 & $--$ \\
    \cellcolor{blue!10}&IRE\textsubscript{RoBERTa-large}~\cite{zhou2021improved} & 46.3\tnote{\dag} & 63.6\tnote{\dag} & 67.0\tnote{\dag} & \underline{74.6} & \underline{83.2} & $--$ \\
    \cellcolor{blue!10}\multirow{-9}{*}{\rotatebox[origin=c]{90}{\textit{Classification-based}}} &RECENT~\cite{lyu-chen-2021-relation} & 40.0\tnote{\dag} & 53.3\tnote{\dag} & 54.2\tnote{\dag} & 67.3\tnote{$\diamondsuit$} & $--$ & $--$ \\
    \midrule
    \cellcolor{blue!10}&NLI\textsubscript{RoBERTa}~\cite{sainz-etal-2021-label} & \underline{56.1} & 64.1 & 67.8 & 71.0 & $--$ & $--$ \\
    \cellcolor{blue!10}&NLI\textsubscript{DeBERTa}~\cite{sainz-etal-2021-label} & \textbf{63.7} & \textbf{69.0} & 67.9 & 73.9 & $--$ & $--$ \\
    \cellcolor{blue!10}\multirow{-3}{*}{\rotatebox[origin=c]{90}{\textit{Ind Sup}}}&KnowPrompt~\cite{chen2021knowprompt} & 51.0\tnote{\dag} & 61.0\tnote{\dag} & 65.2\tnote{\dag} & 72.4 & 82.4 &  \underline{89.6}\tnote{$\diamondsuit$} \\
    \midrule
    \cellcolor{blue!10}&\modelnamens\textsubscript{BART-large}  & 43.6 & 63.8 & 67.9 & 73.3 & 79.2 & 86.3\\
    \cellcolor{blue!10}&\modelnamens\textsubscript{BART-large-cnn} & 50.4 & \underline{65.3} & \underline{68.7} & 73.6 &  81.0 & \underline{89.6}\\
    \cellcolor{blue!10}&\modelnamens\textsubscript{BART-large-xsum} & 50.3 & 64.3 & 68.0 & 73.3 &  81.0 & 89.1 \\
    \cellcolor{blue!10}\multirow{-4}{*}{\rotatebox[origin=c]{90}{\textit{Proposed}}}&\modelnamens\textsubscript{PEGASUS-large}  & 52.0 & 64.9 & \textbf{70.7} & \textbf{75.1} & \textbf{83.3} & \textbf{89.7} \\
    \bottomrule
    \end{tabular}
    }
    \begin{tablenotes}
    \setlength\itemsep{-3.5pt}
    \raggedright
    \item[\dag] {\footnotesize indicates models we re-implement using their official code under the same low-resource setting}.
    \item[\ddag] {\footnotesize indicates results collected from \citet{sainz-etal-2021-label}.}
    \item[$\diamondsuit$] {\footnotesize indicates we reproduce the baseline results (\Cref{ssec:hyperparameter})}.
    \end{tablenotes}
    \end{threeparttable}
    \vspace{-0.2em}
    \caption{Result of \modelname compared with existing methods under both low resource on TACRED and full training scenarios on TACRED, TACREV and SemEval.
    We report micro F1 on TACRED and TACREV, and report macro F1 on SemEval. 
    Baseline F1 scores on the table without special tags are reported by their original studies. 
    Hyphens indicate unavailable results in prior studies. 
    We report \modelname performance with entity information verbalization for consistency. However, \modelname achieves better performance (83.5\%) with mix technique of entity information verbalization and entity typed marker on TACREV (\Cref{tab:result-input-formulation-pegasus-large}).
    We run our models with three different seeds and report the median.
    The best scores are identified with \textbf{bold} and the second best scores are \underline{underlined}.}
    \label{tab:result-existing-methods}
    % \vspace{-0.5em}
\end{table*}

\xhdr{Baselines} We compare \modelname with 8 recent classification-based RE methods: 
(1) \textbf{SpanBERT}~\cite{joshi-etal-2020-spanbert} is a pre-training method designed to better represent and predict spans of text;
(2) \textbf{KnowBERT}~\cite{peters-etal-2019-knowledge} is a PLM embedded multiple KBs;
(3) \textbf{R-BERT}~\cite{wu2019enriching} uses a PLM to encode processed sentences where subject and object entities are marked out;
(4) \textbf{MTB}~\cite{baldini-soares-etal-2019-matching} builds task-agnostic relation representations solely from entity-linked text;
(5) \textbf{K-Adapter}~\cite{wang2021k} infuses knowledge into pretrained language models with adapters.
(6) \textbf{LUKE}~\cite{yamada-etal-2020-luke} modifies the PLM with an entity-aware self-attention mechanism;
(7) \textbf{IRE\textsubscript{RoBERTa-large}}~\cite{zhou2021improved} is an improved baseline model incorporated with typed entity markers;
(8) \textbf{RECENT}~\cite{lyu-chen-2021-relation} introduces type constraint (\Cref{ssec:inference}) and achieve state of the art performance on TACRED.
%mutual restrictions of relations and entity types.
We also compare \modelname with two indirect supervision methods, i.e.
(9) \textbf{NLI\textsubscript{DeBERTa}}~\cite{sainz-etal-2021-label} that formulates RE as NLI, and
(10) \textbf{KnowPrompt}~\cite{chen2021knowprompt} that formulates RE as prompt tuning.

\begin{table}[t]
    \centering
    \begin{threeparttable}
    \small
    \setlength{\tabcolsep}{4.5pt}
    \begin{tabular}{ccccccc}
    \toprule
    \multirow{2}{*}{Template} & 
    \multicolumn{5}{c}{TACRED} &  TACREV \\
    \cline{2-7}
     & 0\% & 1\% & 5\% & 10\% & full & full \\
    \midrule
    \textsc{Semantic1} & 20.6 & \textbf{52.0} & 64.9 & \textbf{70.7} & \textbf{75.0} & \textbf{83.3} \\ 
     \textsc{Semantic2} & 18.5 & 49.5 & \textbf{66.9} & 69.6 & 73.5 & 82.0 \\ 
     \textsc{Structural} & 18.5 & 46.8 & 61.8 & 69.1 & 74.4 & 82.2 \\ 
    \bottomrule
    \end{tabular}
    \end{threeparttable}
    \caption{Analysis of different template designs. The highest scores are highlighted with bold formation. These experiments are conducted with the entity information verbalization technique.}
    \label{tab:result-template-design}
    % \vspace{-0.8em}
\end{table}

\begin{table*}[t]%[htbp]
    \centering
    \begin{threeparttable}
    \small
    \begin{tabular}{cccccc}
    \toprule
    Dataset & \multicolumn{4}{c}{TACRED} & TACREV \\
    \midrule
    Split & 1\% & 5\% & 10\% & full & full\\
    \midrule
    Entity information verbalization    & \textbf{52.0} & \textbf{64.9} & \textbf{70.7} & \textbf{75.1} &  83.3 \\ 
    Entity typed marker (punct)
    & 46.3 & 57.5 & 59.0 & 73.3 &  80.4\\
    Entity information verbalization + Entity typed marker (punct)       
    & 47.6 & 60.2 & 67.8 & 75.0 &  \textbf{83.5} \\ 
    \bottomrule
    \end{tabular}
    % \vspace{-0.5em}
    \end{threeparttable}
    \caption{Analysis of different input formulation techniques on \emph{PEGASUS-large}}
    \label{tab:result-input-formulation-pegasus-large}
    % \vspace{-1em}
\end{table*}

\xhdr{Low resource setting} We evaluate the performance of \modelname under low-resource scenarios. To do so, we use the same splits of \citet{sainz-etal-2021-label} to build the TACRED datasets with 1/5/10 percent of training and development samples.

\xhdr{Model configurations} We develop \modelname based on two widely pretrained generative models BART~\cite{lewis-etal-2020-bart} and PEGASUS~\cite{zhang2020pegasus}. BART is a denoising autoencoder
for pretraining sequence-to-sequence models. We use two summarization models \emph{BART-large-cnn} and \emph{BART-large-xsum} that are finetuned with CNN/Dailymail~\cite{hermann2015teaching} and XSum~\cite{narayan-etal-2021-planning}, respectively. 
We also consider \emph{BART-large} as a baseline without indirect supervision of summarization. PEGASUS is a sequence-to-sequence model pretrained with a gap sentences generation task, which significantly benefits various summarization downstream tasks. Similarly, we use \emph{PEGASUS-large} as a stronger initial checkpoint than BARTs. We use grid search to find optimal hyperparameters for finetuning summarization models. The best hyperparameters of our experiments and re-implementation of baselines are shown in \Cref{ssec:hyperparameter}.

\subsection{Results}\label{ssec:results}

We present our main results on both full training and low-resource settings in \Cref{tab:result-existing-methods}. 
We report the performance of \modelname with entity information verbalization technique, which is proved to be the best way of input sequence construction (\Cref{ssec:conversion}) in most settings as shown in our ablation study (\Cref{ssec:input_formulation}).

\xhdr{Performance comparison}
With 1\%, 5\%, or 10\% training data of TACRED, \modelname with summarization backbones (\modelnamens\textsubscript{BART-large-cnn}, \modelnamens\textsubscript{BART-large-xsum} and \modelnamens\textsubscript{PEGASUS-large}) and other baselines with indirect supervision consistently outperform classification-based RE models except IRE, which indicates the benefit of indirect supervision.
Although NLI\textsubscript{DeBERTa} significantly outperforms other methods with 1\% and 5\% training data, \modelname has significant improvement on 10\% TACRED, which outperform NLI\textsubscript{DeBERTa} for 2.8\% and KnowPrompt for 5.5\%. Furthermore, the performance of \modelname is 1.2\% higher than that of NLI\textsubscript{DeBERTa}, and 2.7\% higher than that of KnowPrompt in full training setting of TACRED, which suggests that \modelname achieves the best performance with adequate training samples among all indirect supervision baselines.
\modelname also achieves the best performance among all baselines, and exceeds the second best model IRE\textsubscript{RoBERTa-large} by 0.5\% in F1.
Besides, \modelname also achieves the best F1 score on TACREV. 
And it outperforms all classification-based models on SemEval, while its performance is comparable to KnowPrompt.

\xhdr{Effectiveness of indirect supervision}
We further evaluate \modelname based on different pretrained models, as shown in~\Cref{tab:result-existing-methods}.
We observe that models finetuned on summarization tasks~(CNN and XSum) generally lead to better performance, especially in the low-resource setting.
For example, \modelnamens\textsubscript{BART-large-cnn} outperforms \modelnamens\textsubscript{BART-large} by 6.8\% on the 1\% split of TACRED, while this improvement diminishes to 0.3\% on the full dataset.
Besides, pretrained models that perform better on summarization tasks also indicate better performance on RE.
Particularly, \modelname based on \emph{PEGASUS-large}, which outperforms \emph{BART-large} on summarization tasks, outperforms all other versions under both low-resource and full-dataset setting.
Both observations show a strong correlation between the performance in summarization and RE, indicating that indirect supervision from summarization is beneficial for RE models.

\subsection{Ablation Study}\label{ssec:ablation}

\begin{table*}[h]
    \centering
    \small
    \begin{threeparttable}
    \setlength{\tabcolsep}{5pt}
    \resizebox{\textwidth}{!}{
    \begin{tabular}{c|c|cccc}
    \toprule
    Technique & Example & M\tnote{1} & S\tnote{1} & T\tnote{1} & TS\tnote{1} \\
    \midrule
    Entity information verbalization & \tabincell{l}{\texttt{Context}\tnote{2} . ... \{subj\} ... \{obj\} ...} & \Checkmark & \XSolidBrush & \Checkmark & \Checkmark\\
    % \hline
    \midrule
    \tabincell{c}{Entity marker \\\cite{zhang-etal-2019-ernie}} & \tabincell{l}{... <e1> \{subj\} </e1> ... <e2> \{obj\} </e2> ...} & \Checkmark & \Checkmark & \XSolidBrush & \XSolidBrush\\
    % \hline
    \tabincell{c}{Entity typed marker \\\cite{zhong-chen-2021-frustratingly}} & \tabincell{l}{... <e1-\{subj-type\}> \{subj\} </e1-\{subj-type\}> \\... <e2-\{obj-type\}> \{obj\} </e2-\{obj-type\}> ...} & \Checkmark & \Checkmark & \Checkmark & \XSolidBrush\\
    % \hline
    \tabincell{c}{Entity typed marker (punct) \\\cite{zhou2021improved}} & \tabincell{l}{... @ * \{subj-type\} * \{subj\} @ ... \# $\wedge$ \{obj-type\} $\wedge$ \{obj\} \# ...} & \Checkmark & \Checkmark & \Checkmark & \Checkmark \\
    % \hline
    \midrule
    \tabincell{c}{Entity information verbalization + Entity typed marker} & \tabincell{l}{\texttt{Context}\tnote{2} ... @ * \{subj-type\} * \{subj\} @ \\... \# $\wedge$ \{obj-type\} $\wedge$ \{obj\} \# ...} & \Checkmark & \Checkmark & \Checkmark & \Checkmark \\
    % \hline
    \tabincell{c}{Entity information verbalization + Entity typed marker (punct)} & \tabincell{l}{\texttt{Context}\tnote{2} ... @ * \{subj-type\} * \{subj\} @ \\... \# $\wedge$ \{obj-type\} $\wedge$ \{obj\} \# ...} & \Checkmark & \Checkmark & \Checkmark & \Checkmark \\
    \bottomrule
    \end{tabular}
    }
    \begin{tablenotes}
        \footnotesize 
        \item[1] Column names are short for mentions, spans, types and type semantics, respectively.
        \item[2] Augmented context are generated with the template: ``\emph{The \{subj\} entity is \{subj\} .\\ The \{obj\} entity is \{obj\} . The type of \{subj\} is \{subj-type\} . The type of \{obj\} is \{obj-type\} }''
      \end{tablenotes}
    \end{threeparttable}
    \caption{Sentence processing techniques for incorporating entity information.}
    \label{tab:entity-information-injection-techniques}
\end{table*}

\begin{table*}[h]
    \centering
    \begin{threeparttable}
    \small
    \begin{tabular}{ccccccc}
    \toprule
    \multirow{2}{*}{Technique} & \multicolumn{3}{c}{TACRED} & \multicolumn{3}{c}{TACREV} \\
    \cline{2-7}
    & P & R & F1 & P & R & F1 \\
    \midrule
    Entity information verbalization & 71.8 & 75.1 & 73.4 & 78.4 & 83.8 & 81.0 \\ 
    Entity tag & 71.3 & 71.0 & 71.1 & 81.0 & 75.2 & 78.0 \\ 
    Entity typed tag & 73.5 & 69.7 & 71.5 & 79.9 & 79.6 & 79.8 \\ 
    Entity typed tag(punct) & 70.3 & 74.0 & 72.1 & 81.1 & 79.3 & 80.2 \\ 
    Entity infromation verbalization + Entity tag & 73.4 & 71.8 & 72.6 & 78.7 & 81.7 & 80.2 \\ 
    Entity infromation verbalization + Entity typed tag        & 70.6 & 73.5 & 72.1 & 82.8 & 80.0 & \textbf{81.4} \\
    Entity infromation verbalization + Entity typed tag(punct) & 72.6 & 74.5 & \textbf{73.6} & 82.1 & 79.8 & 81.0 \\
    \bottomrule
    \end{tabular}
    \end{threeparttable}
    \caption{Analysis of different input formulation techniques on \emph{bart-large-cnn}. We report micro F1 scores on both datasets. The best F1 score is identified with \textbf{bold}.}
    \label{tab:result-input-formulation-bart-large-cnn}
\end{table*}

We provide the following analyses to further evaluate core components of \modelname, including different template designs, sentence conversion techniques and Trie scoring. We also report the ablation study on type constrained inference and calibration of NA relation in \Cref{ssec:type_constrained_decoding,ssec:analysis_of_calibration}, respectively.

\xhdr{Relation template design}\label{ssec:template_design}
Template design is a manual part of \modelname. The semantic meaning of a relation can be verbalized in different ways, leading to varied performance. 
In \Cref{tab:result-template-design}, we compare three representative types of templates with \emph{pegasus-large} in this ablation study to show how verbalization templates influence the performance of \modelname.
\textsc{Semantic1} thereof denotes semantic templates beginning with subject entities and ending with object entities, which is showed in Appx. \Cref{tab:semantic1-templates}. \textsc{Semantic2} are also semantic templates that intuitively describe the relation between two entiteis with a pattern ``\emph{The relation between \{subj\} and \{obj\} is \{relation\}}'', which is showed in Appx. \Cref{tab:semantic2-templates}.
\textsc{Structural} marks structural templates forming in a triplet structure ``\{subj\} \{relation\} \{obj\}'', which is showed in Appx. \Cref{tab:structural-templates}.
Furthermore, we also set up a zero-shot setting where the model directly infers on the test set of TACRED after calibration on the development set.
The results from different templates are reported on both low-resource and full-training scenarios. 
First of all, we observe that the two semantic templates consistently outperform structural templates, indicating that semantic templates are more suitable for acquiring indirect supervision from summarization.
Besides, comparing two semantic templates, we find that \emph{Semantic1} works better with \emph{pegasus-large}, which suggests that the optimal verbalization may vary.
And this difference is more obvious under low resource scenarios.
Consequently, zero-shot inference is an effective and efficient method for evaluating manual-designed templates. In future work, we can investigate how to improve this part by prompt tuning.

\xhdr{Input conversion}\label{ssec:input_formulation}
We conduct experiments to evaluate various input sentence conversion techniques for injecting entity information into source sentences (\Cref{ssec:conversion}). 
We first conduct experiments on six different input formulations on \emph{bart-large-cnn}, which is listed in \Cref{tab:entity-information-injection-techniques} and results are shown in Appx. \Cref{tab:result-input-formulation-bart-large-cnn}. This experiment indicates entity typed marker with punctuation is the best technique for \modelname among all entity marker techniques. Then, we further evaluate three techniques on \emph{pegasus-large} under both full training and low resource scenarios. \Cref{tab:result-input-formulation-pegasus-large} shows entity information verbalization achieves significantly better performance under low resource scenarios compared with marker and mix techniques. This is because entity information verbalization transforms input to better fit the input of summarization, while additional markers need more data to learn their representations. In the full training setting, the mixing technique marginally outperforms entity information verbalization.

\xhdr{Trie scoring} Trie scoring uses teacher forcing to constraint models focusing on candidate templates and have the advantage of efficiency compared with directly comparing likelihoods of all templates. Furthermore, we also make comparisons between Trie scoring and two basic scoring methods on full TACRED with \modelnamens\textsubscript{pegasus-large}. The first one is to generate the summary of an example and uses ROUGE-L \cite{lin-2004-rouge} scores between summary and candidate templates as prediction scores. This method achieves 74.7\% on TACRED, which is 0.4\% less than that of Trie scoring. Another method is adding a copy mechanism \cite{zeng-etal-2018-extracting} to ensure summaries begin with head entities, which achieves a comparable performance of the previous method (74.8\%), which further proves the advantages of Trie scoring.

\section{Conclusion}
We propose a new method \modelname that leverages indirect supervision from summarization tasks to improve RE.
To do so, we verbalize relations with semantic templates, and augments entity information as parts of the linguistic context in the inputs to allow them to suit the formation of summarization. We also incorporate \modelname with constrained inference based on Trie scoring, as well as inference with abstention and entity type constraints. Extensive experiments show that such indirectly supervised RE by \modelname lead to more precise and resource-efficient RE. Future work includes further developing our model on document-level RE tasks and minimizing manual effort in template design with prompt tuning \cite{li-liang-2021-prefix}.

\section*{Acknowledgement}

We appreciate the reviewers for their insightful
comments and suggestions. This work is partly
supported by the National Science Foundation of
United States Grant IIS 2105329, and a Cisco Faculty Research Award.

%\clearpage
\section*{Limitations}
\modelname assumes that %the annotation of summarization 
summarization data
and manual-designed verbalization templates
of relations are easy to obtain.
This assumption is
hold in the %most common domains.
general domain.
However, %they can be probably
%counterfactual
obtaining such data and templates can still be difficult in specific lower-resource domains.
For example, summarization data in other languages are not as rich as those in English. 
Hence, \modelname may benefit less from indirect supervision signals when it is adapted to multilingual scenarios. Besides, designing templates in specific domains, such as biomedical
relation extraction, may require extra involvement of expert knowledge. %which will increase manual efforts in \modelname. 
Although we put certain manual efforts in template design, automatically optimizing templates are also feasible for \modelname and can be explored in future work, as described in the Conclusion section.
%\clearpage

%\clearpage
%\section{Author response}
%\input{response.tex}
%\clearpage

\bibliography{anthology,custom}

\clearpage
\appendix

\section{Appendix}
\label{sec:appendix}
\subsection{Dataset Statistics}

Statistics of the RE datasets are listed in \Cref{tab:dataset-statistics}.

\begin{table}[h]
    \centering
    \small
    \begin{threeparttable}
    \resizebox{\linewidth}{!}{
    \begin{tabular}{ccccc}
    \toprule
    Dataset & \#Train & \#Dev & \#Test & \#Relation \\
    \midrule
    SemEval & 8000 & - & 2717 & 19 \\
    TACRED & 68124 & 22631 & 15509 & 42 \\
    TACREV & 68124 & 22631 & 15509 & 42 \\
    \midrule
    TACRED(1\%) & 682 & 227 & \multirow{3}{*}{15509} & \multirow{3}{*}{42}\\
    TACRED(5\%) & 3407 & 1133 & & \\
    TACRED(10\%) & 6815 & 2265 & & \\
    \bottomrule
    \end{tabular}
    }
    \end{threeparttable}
    % \vspace{-0.5em}
    \caption{Statistics of datasets}
    \label{tab:dataset-statistics}
    % \vspace{-1em}
\end{table}

\subsection{Analysis of type constrained decoding} \label{ssec:type_constrained_decoding}

In this ablation study, we make comparisons between \modelname with and without typed constrained decoding. The results is demonstrated in \Cref{tab:result-type-constraint}. Type constraint brings improvement for 0.2\% in average in TACRED and has comparable performance on TACREV. Type-relation mapping is inherently involved in training data, so this ablation study proves \modelname can learn type-relation mapping from data.

\begin{table}[h]
    \centering
    \begin{threeparttable}
    \small
    \begin{tabular}{ccc}
    \toprule
    Dataset & TACRED & TACREV\\
    \midrule
    bart-large-cnn & 73.6 & 81.0 \\
    - type constraint & 73.4 & 81.0 \\
    \midrule
    bart-large-xsum & 73.3 & 81.0 \\
    - type constraint & 73.1 & 81.0 \\
    \midrule
    pegasus-large & 75.1 & 83.3 \\
    - type constraint & 75.0 & 83.3\\
    \midrule
    average gap & -0.2 & 0\\
    \bottomrule
    \end{tabular}
    \end{threeparttable}
    \caption{Comparison between \modelname with and without type constraint decoding. We report micro F1 on TACRED and TACREV.}
    \label{tab:result-type-constraint}
\end{table}

\subsection{Analysis of calibration for NA relation} \label{ssec:analysis_of_calibration}

In this ablation study, we make comparisons between \modelname with and without calibration for NA relation. The results is demonstrated in \Cref{tab:analysis_of_calibration}. Calibration brings improvement for 0.3\% in average on TACRED and 0.1\% in average on TACREV.

\begin{table}[h]
    \centering
    \begin{threeparttable}
    \small
    \begin{tabular}{ccc}
    \toprule
    Dataset & TACRED & TACREV \\
    \midrule
    bart-large-cnn & 73.6 & 81.0 \\
    - calibration & 73.2 & 80.9 \\
    \midrule
    bart-large-xsum & 73.3 & 81.0 \\
    - calibration & 72.9  & 80.9 \\
    \midrule
    pegasus-large & 75.1 & 83.3\\
    - calibration & 74.8 & 83.2\\
    \midrule
    average gap &  -0.3 & -0.1 \\
    \bottomrule
    \end{tabular}
    \end{threeparttable}
    \caption{Comparison between \modelname with and without calibration for NA relation We report micro F1 on TACRED and TACREV.}
    \label{tab:analysis_of_calibration}
\end{table}

\subsection{Hyper-parameters and reimplementation}\label{ssec:hyperparameter}

This section details the training and inference processes of baselines and our models. We train and inference all models with PyTorch and Huggingface Transformers on GeForce RTX 2080 or NVIDIA RTX A5000 GPUs. All optimization uses Adam and linear scheduler. A weight decay is used for regularization. We run all experiments on three seeds $[0, 100, 500]$ and report the median.
Specifically, the best hyperparameters for full training setting with \emph{pegasus-large} are listed below:

\begin{itemize}[leftmargin=1em]
    \setlength\itemsep{0em}
    \item learning rate: 1e-4
    \item weight decay: 5e-6
    \item epoch number: 20
    \item max source length: 256
    \item max target length: 64
    \item gradient accumulation steps: 2
    \item warm up steps: 1000
\end{itemize}

The best hyperparameters for low-resource setting with \emph{pegasus-large} are listed below:

\begin{itemize}[leftmargin=1em]
    \setlength\itemsep{0em}
    \item learning rate: 1e-5
    \item weight decay: 5e-6
    \item epoch number: 100
    \item max source length: 256
    \item max target length: 64
    \item gradient accumulation steps: 2
    \item warm up steps: 0
\end{itemize}

With the 1/5/10\% indices of TACRED provided by \cite{sainz-etal-2021-label}\footnote{Github repository of low-resource indices: \url{https://github.com/osainz59/Ask2Transformers}}, we re-implement RECENT~\cite{lyu-chen-2021-relation} and test it under both low-resource and full-training scenarios\footnote{Github repository of RECENT: \url{https://github.com/Saintfe/RECENT}}. However, we find the origin evaluation scripts provided by the author has a serious issue which wrongly corrects all false positive samples of the binary classifier as true negative samples. So the recall of NA is always 100\% and precision of positive relations is unreasonably high. We correct this issue and the test results significantly differ from origin results reported by previous study. We also test IRE~\cite{zhou2021improved} and KnowPrompt~\cite{chen2021knowprompt} on low resource datasets and search the best hyperparameter with grid searching\footnote{Github repository of IRE: \url{https://github.com/wzhouad/RE\_improved\_baseline}}\footnote{Github repository of KnowPrompt: \url{https://github.com/zjunlp/KnowPrompt}}. The previous work of KnowPrompt reports the micro F1 score on SemEval. We train KnowPrompt on SemEval with codes and hyper-parameters provided by authors and re-evaluate it with official macro-F1 scoring method.

%\subsection{Additional information of input conversion techniques}\label{ssec:additional_input_conversion}

%We hereby list all input conversion techniques we experiment in this work in \Cref{tab:entity-information-injection-techniques}. \Cref{tab:result-input-formulation-bart-large-cnn} shows additional results on the  \emph{bart-large-cnn} model, which provides the same conclusion as results on \emph{pegasus-large}.

\begin{table*}[h]
    \centering
    \small
    \begin{threeparttable}
    \setlength{\tabcolsep}{5pt}
    \resizebox{\textwidth}{!}{
    \begin{tabular}{c|c|cccc}
    \toprule
    Technique & Example & M\tnote{1} & S\tnote{1} & T\tnote{1} & TS\tnote{1} \\
    \midrule
    Entity information verbalization & \tabincell{l}{\texttt{Context}\tnote{2} . ... \{subj\} ... \{obj\} ...} & \Checkmark & \XSolidBrush & \Checkmark & \Checkmark\\
    % \hline
    \midrule
    \tabincell{c}{Entity marker \\\cite{zhang-etal-2019-ernie}} & \tabincell{l}{... <e1> \{subj\} </e1> ... <e2> \{obj\} </e2> ...} & \Checkmark & \Checkmark & \XSolidBrush & \XSolidBrush\\
    % \hline
    \tabincell{c}{Entity typed marker \\\cite{zhong-chen-2021-frustratingly}} & \tabincell{l}{... <e1-\{subj-type\}> \{subj\} </e1-\{subj-type\}> \\... <e2-\{obj-type\}> \{obj\} </e2-\{obj-type\}> ...} & \Checkmark & \Checkmark & \Checkmark & \XSolidBrush\\
    % \hline
    \tabincell{c}{Entity typed marker (punct) \\\cite{zhou2021improved}} & \tabincell{l}{... @ * \{subj-type\} * \{subj\} @ ... \# $\wedge$ \{obj-type\} $\wedge$ \{obj\} \# ...} & \Checkmark & \Checkmark & \Checkmark & \Checkmark \\
    % \hline
    \midrule
    \tabincell{c}{Entity information verbalization + Entity typed marker} & \tabincell{l}{\texttt{Context}\tnote{2} ... @ * \{subj-type\} * \{subj\} @ \\... \# $\wedge$ \{obj-type\} $\wedge$ \{obj\} \# ...} & \Checkmark & \Checkmark & \Checkmark & \Checkmark \\
    % \hline
    \tabincell{c}{Entity information verbalization + Entity typed marker (punct)} & \tabincell{l}{\texttt{Context}\tnote{2} ... @ * \{subj-type\} * \{subj\} @ \\... \# $\wedge$ \{obj-type\} $\wedge$ \{obj\} \# ...} & \Checkmark & \Checkmark & \Checkmark & \Checkmark \\
    \bottomrule
    \end{tabular}
    }
    \begin{tablenotes}
        \footnotesize 
        \item[1] Column names are short for mentions, spans, types and type semantics, respectively.
        \item[2] Augmented context are generated with the template: ``\emph{The \{subj\} entity is \{subj\} .\\ The \{obj\} entity is \{obj\} . The type of \{subj\} is \{subj-type\} . The type of \{obj\} is \{obj-type\} }''
      \end{tablenotes}
    \end{threeparttable}
    \caption{Sentence processing techniques for incorporating entity information.}
    \label{tab:entity-information-injection-techniques}
\end{table*}

\begin{table*}[h]
    \centering
    \begin{threeparttable}
    \small
    \begin{tabular}{ccccccc}
    \toprule
    \multirow{2}{*}{Technique} & \multicolumn{3}{c}{TACRED} & \multicolumn{3}{c}{TACREV} \\
    \cline{2-7}
    & P & R & F1 & P & R & F1 \\
    \midrule
    Entity information verbalization & 71.8 & 75.1 & 73.4 & 78.4 & 83.8 & 81.0 \\ 
    Entity tag & 71.3 & 71.0 & 71.1 & 81.0 & 75.2 & 78.0 \\ 
    Entity typed tag & 73.5 & 69.7 & 71.5 & 79.9 & 79.6 & 79.8 \\ 
    Entity typed tag(punct) & 70.3 & 74.0 & 72.1 & 81.1 & 79.3 & 80.2 \\ 
    Entity infromation verbalization + Entity tag & 73.4 & 71.8 & 72.6 & 78.7 & 81.7 & 80.2 \\ 
    Entity infromation verbalization + Entity typed tag        & 70.6 & 73.5 & 72.1 & 82.8 & 80.0 & \textbf{81.4} \\
    Entity infromation verbalization + Entity typed tag(punct) & 72.6 & 74.5 & \textbf{73.6} & 82.1 & 79.8 & 81.0 \\
    \bottomrule
    \end{tabular}
    \end{threeparttable}
    \caption{Analysis of different input formulation techniques on \emph{bart-large-cnn}. We report micro F1 scores on both datasets. The best F1 score is identified with \textbf{bold}.}
    \label{tab:result-input-formulation-bart-large-cnn}
\end{table*}

\subsection{Manual-constructed templates}

In this subsection, we display our manual-constructed templates for SemEval (\Cref{tab:semantic1-templates-semeval}), and three templates designed for TACRED, which are \textit{Semantic1} (\Cref{tab:semantic1-templates}), \textit{Semantic2}(\Cref{tab:semantic2-templates}), and \textit{Structural}(\Cref{tab:structural-templates}).

\begin{table*}[h]
    \centering
    \begin{threeparttable}
    \small
    \begin{tabular}{cc}
    \toprule
    Relation & Template \\
    \midrule
   org:country\_of\_headquarters & \{subj\} has a headquarter in the country \{obj\}\\
org:parents & \{subj\} has the parent company \{obj\}\\
per:stateorprovince\_of\_birth & \{subj\} was born in the state or province \{obj\}\\
per:spouse & \{subj\} is the spouse of \{obj\}\\
per:origin & \{subj\} has the nationality \{obj\}\\
per:date\_of\_birth & \{subj\} has birthday on \{obj\}\\
per:schools\_attended & \{subj\} studied in \{obj\}\\
org:members & \{subj\} has the member \{obj\}\\
org:founded & \{subj\} was founded in \{obj\}\\
per:stateorprovinces\_of\_residence & \{subj\} lives in the state or province \{obj\}\\
per:date\_of\_death & \{subj\} died in the date \{obj\}\\
org:shareholders & \{subj\} has shares hold in \{obj\}\\
org:website & \{subj\} has the website \{obj\}\\
org:subsidiaries & \{subj\} owns \{obj\}\\
per:charges & \{subj\} is convicted of \{obj\}\\
org:dissolved & \{subj\} dissolved in \{obj\}\\
org:stateorprovince\_of\_headquarters & \{subj\} has a headquarter in the state or province \{obj\}\\
per:country\_of\_birth & \{subj\} was born in the country \{obj\}\\
per:siblings & \{subj\} is the siblings of \{obj\}\\
org:top\_members/employees & \{subj\} has the high level member \{obj\}\\
per:cause\_of\_death & \{subj\} died because of \{obj\}\\
per:alternate\_names & \{subj\} has the alternate name \{obj\}\\
org:number\_of\_employees/members & \{subj\} has the number of employees \{obj\}\\
per:cities\_of\_residence & \{subj\} lives in the city \{obj\}\\
org:city\_of\_headquarters & \{subj\} has a headquarter in the city \{obj\}\\
per:children & \{subj\} is the parent of \{obj\}\\
per:employee\_of & \{subj\} is the employee of \{obj\}\\
org:political/religious\_affiliation & \{subj\} has political affiliation with \{obj\}\\
per:parents & \{subj\} has the parent \{obj\}\\
per:city\_of\_birth & \{subj\} was born in the city \{obj\}\\
per:age & \{subj\} has the age \{obj\}\\
per:countries\_of\_residence & \{subj\} lives in the country \{obj\}\\
org:alternate\_names & \{subj\} is also known as \{obj\}\\
per:religion & \{subj\} has the religion \{obj\}\\
per:city\_of\_death & \{subj\} died in the city \{obj\}\\
per:country\_of\_death & \{subj\} died in the country \{obj\}\\
org:founded\_by & \{subj\} was founded by \{obj\}"\\
    \bottomrule
    \end{tabular}
    \end{threeparttable}
    \caption{First semantic templates for TACRED, where \{subj\} and \{obj\} are the placeholders for subject and object entities.}
    \label{tab:semantic1-templates}
\end{table*}

\begin{table*}[h]
    \centering
    \begin{threeparttable}
    \small
    \begin{tabular}{cc}
    \toprule
    Relation & Template \\
    \midrule
    no\_relation &     The relation between \{subj\} and \{obj\} is not available\\
    per:stateorprovince\_of\_death &     The relation between \{subj\} and \{obj\} is state or province of death\\
    per:title &     The relation between \{subj\} and \{obj\} is title\\
    org:member\_of &     The relation between \{subj\} and \{obj\} is member of\\
    per:other\_family &     The relation between \{subj\} and \{obj\} is other family\\
    org:country\_of\_headquarters &     The relation between \{subj\} and \{obj\} is country of headquarters\\
    org:parents &     The relation between \{subj\} and \{obj\} is parents of the organization\\
    per:stateorprovince\_of\_birth &     The relation between \{subj\} and \{obj\} is state or province of birth\\
    per:spouse &     The relation between \{subj\} and \{obj\} is spouse\\
    per:origin &     The relation between \{subj\} and \{obj\} is origin\\
    per:date\_of\_birth &     The relation between \{subj\} and \{obj\} is date of birth\\
    per:schools\_attended &     The relation between \{subj\} and \{obj\} is schools attended\\
    org:members &     The relation between \{subj\} and \{obj\} is members\\
    org:founded &     The relation between \{subj\} and \{obj\} is founded\\
    per:stateorprovinces\_of\_residence &     The relation between \{subj\} and \{obj\} is state or province of residence\\
    per:date\_of\_death &     The relation between \{subj\} and \{obj\} is date of death\\
    org:shareholders &     The relation between \{subj\} and \{obj\} is shareholders\\
    org:website &     The relation between \{subj\} and \{obj\} is website\\
    org:subsidiaries &     The relation between \{subj\} and \{obj\} is subsidiaries\\
    per:charges &     The relation between \{subj\} and \{obj\} is charges\\
    org:dissolved &     The relation between \{subj\} and \{obj\} is dissolved\\
    org:stateorprovince\_of\_headquarters &     The relation between \{subj\} and \{obj\} is state or province of headquarters\\
    per:country\_of\_birth &     The relation between \{subj\} and \{obj\} is country of birth\\
    per:siblings &     The relation between \{subj\} and \{obj\} is siblings\\
    org:top\_members/employees &     The relation between \{subj\} and \{obj\} is top members or employees\\
    per:cause\_of\_death &     The relation between \{subj\} and \{obj\} is cause of death\\
    per:alternate\_names &     The relation between \{subj\} and \{obj\} is person alternative names\\
    org:number\_of\_employees/members &     The relation between \{subj\} and \{obj\} is number of employees or members\\
    per:cities\_of\_residence &     The relation between \{subj\} and \{obj\} is cities of residence\\
    org:city\_of\_headquarters &     The relation between \{subj\} and \{obj\} is city of headquarters\\
    per:children &     The relation between \{subj\} and \{obj\} is children\\
    per:employee\_of &     The relation between \{subj\} and \{obj\} is employee of\\
    org:political/religious\_affiliation &     The relation between \{subj\} and \{obj\} is political and religious affiliation\\
    per:parents &     The relation between \{subj\} and \{obj\} is parents of the person\\
    per:city\_of\_birth &     The relation between \{subj\} and \{obj\} is city of birth\\
    per:age &     The relation between \{subj\} and \{obj\} is age\\
    per:countries\_of\_residence &     The relation between \{subj\} and \{obj\} is countries of residence\\
    org:alternate\_names &     The relation between \{subj\} and \{obj\} is organization alternate names\\
    per:religion &     The relation between \{subj\} and \{obj\} is religion\\
    per:city\_of\_death &     The relation between \{subj\} and \{obj\} is city of death\\
    per:country\_of\_death &     The relation between \{subj\} and \{obj\} is country of death\\
    org:founded\_by &     The relation between \{subj\} and \{obj\} is founded by"\\
    \bottomrule
    \end{tabular}
    \end{threeparttable}
    \caption{Second semantic templates for TACRED, where \{subj\} and \{obj\} are the placeholders for subject and object entities.}
    \label{tab:semantic2-templates}
\end{table*}

\begin{table*}[h]
    \centering
    \begin{threeparttable}
    \small
    \begin{tabular}{cc}
    \toprule
    Relation & Template \\
    \midrule
    no\_relation & \{subj\} no relation \{obj\} \\
    per:stateorprovince\_of\_death & \{subj\} person state or province of death \{obj\} \\
    per:title & \{subj\} person title \{obj\} \\
    org:member\_of & \{subj\} organization member of \{obj\} \\
    per:other\_family & \{subj\} person other family \{obj\} \\
    org:country\_of\_headquarters & \{subj\} organization country of headquarters \{obj\} \\
    org:parents & \{subj\} organization parents \{obj\} \\
    per:stateorprovince\_of\_birth & \{subj\} person state or province of birth \{obj\} \\
    per:spouse & \{subj\} person spouse \{obj\} \\
    per:origin & \{subj\} person origin \{obj\} \\
    per:date\_of\_birth & \{subj\} person date of birth \{obj\} \\
    per:schools\_attended & \{subj\} person schools attended \{obj\} \\
    org:members & \{subj\} organization members \{obj\} \\
    org:founded & \{subj\} organization founded \{obj\} \\
    per:stateorprovinces\_of\_residence & \{subj\} person state or provinces of residence \{obj\} \\
    per:date\_of\_death & \{subj\} person date of death \{obj\} \\
    org:shareholders & \{subj\} organization shareholders \{obj\} \\
    org:website & \{subj\} organization website \{obj\} \\
    org:subsidiaries & \{subj\} organization subsidiaries \{obj\} \\
    per:charges & \{subj\} person charges \{obj\} \\
    org:dissolved & \{subj\} organization dissolved \{obj\} \\
    org:stateorprovince\_of\_headquarters & \{subj\} organization state or province of headquarters \{obj\} \\
    per:country\_of\_birth & \{subj\} person country of birth \{obj\} \\
    per:siblings & \{subj\} person siblings \{obj\} \\
    org:top\_members/employees & \{subj\} organization top members or employees \{obj\} \\
    per:cause\_of\_death & \{subj\} person cause of death \{obj\} \\
    per:alternate\_names & \{subj\} person alternate names \{obj\} \\
    org:number\_of\_employees/members & \{subj\} organization number of employees or members \{obj\} \\
    per:cities\_of\_residence & \{subj\} person cities of residence \{obj\} \\
    org:city\_of\_headquarters & \{subj\} organization city of headquarters \{obj\} \\
    per:children & \{subj\} person children \{obj\} \\
    per:employee\_of & \{subj\} person employee of \{obj\} \\
    org:political/religious\_affiliation & \{subj\} organization political or religious affiliation \{obj\} \\
    per:parents & \{subj\} person parents \{obj\} \\
    per:city\_of\_birth & \{subj\} person city of birth \{obj\} \\
    per:age & \{subj\} person age \{obj\} \\
    per:countries\_of\_residence & \{subj\} person countries of residence \{obj\} \\
    org:alternate\_names & \{subj\} organization alternate names \{obj\} \\
    per:religion & \{subj\} person religion \{obj\} \\
    per:city\_of\_death & \{subj\} person city of death \{obj\} \\
    per:country\_of\_death & \{subj\} person country of death \{obj\} \\
    org:founded\_by & \{subj\} organization founded by \{obj\} \\
    \bottomrule
    \end{tabular}
    \end{threeparttable}
    \caption{Structural templates for TACRED, where \{subj\} and \{obj\} are the placeholders for subject and object entities.}
    \label{tab:structural-templates}
\end{table*}

\begin{table*}[h]
    \centering
    \begin{threeparttable}
    \small
    \begin{tabular}{cc}
    \toprule
    Relation & Template \\
    \midrule
    Other & \{subj\} is not related to \{obj\}\\
    Component-Whole(e1,e2) & \{subj\} is the component of \{obj\}\\
    Component-Whole(e2,e1) & \{subj\} has the component \{obj\}\\
    Instrument-Agency(e1,e2) & \{subj\} is the instrument of \{obj\}\\
    Instrument-Agency(e2,e1) & \{subj\} has the instrument \{obj\}\\
    Member-Collection(e1,e2) & \{subj\} is the member of \{obj\}\\
    Member-Collection(e2,e1) & \{subj\} has the member \{obj\}\\
    Cause-Effect(e1,e2) & \{subj\} has the effect \{obj\}\\
    Cause-Effect(e2,e1) & \{subj\} is the effect of \{obj\}\\
    Entity-Destination(e1,e2) & \{subj\} locates in \{obj\}\\
    Entity-Destination(e2,e1) & \{subj\} is the destination of \{obj\}\\
    Content-Container(e1,e2) & \{subj\} is the content in \{obj\}\\
    Content-Container(e2,e1) & \{subj\} contains \{obj\}\\
    Message-Topic(e1,e2) & \{subj\} is the message on \{obj\}\\
    Message-Topic(e2,e1) & \{subj\} is the topic of \{obj\}\\
    Product-Producer(e1,e2) & \{subj\} is the product of \{obj\}\\
    Product-Producer(e2,e1) & \{subj\} produces \{obj\}\\
    Entity-Origin(e1,e2) & \{subj\} origins from \{obj\}\\
    Entity-Origin(e2,e1) & \{subj\} is the origin of \{obj\}\\
 \bottomrule
    \end{tabular}
    \end{threeparttable}
    \caption{Semantic templates for SemEval, where \{subj\} and \{obj\} are the placeholders for subject and object entities.}
    \label{tab:semantic1-templates-semeval}
\end{table*}

\begin{comment}
\subsection{Limitations and Potential Risks}

\xhdr{Limitations} \modelname assumes that the annotation of summarization and manual-designed templates of relations are cheap to obtain. This assumption is hold in the most common domains. However, they can be probably counterfactual in specific domains. For eaxmple, designing templates for biomedical relation extraction needs expert knowledge, which will increase manual efforts in \modelname.

\xhdr{Potential risks} \modelname are based on pretrained models, such as BARTs~\cite{lewis-etal-2020-bart}. Therefore, the summarization of our models can be influenced by the training corpus of these pretrained models, which may raise potential risks of generating malicious, counterfactual, and biased sentences and causing ethics concerns. We suggest examing these issues before applying \modelname in real-world.
\end{comment}

\end{document}